\documentclass{article}

\usepackage{arxiv}

\usepackage[utf8]{inputenc} % allow utf-8 input
\usepackage[T1]{fontenc}    % use 8-bit T1 fonts
\usepackage{hyperref}       % hyperlinks
\usepackage{url}            % simple URL typesetting
\usepackage{booktabs}       % professional-quality tables
\usepackage{amsfonts}       % blackboard math symbols
\usepackage{nicefrac}       % compact symbols for 1/2, etc.
\usepackage{microtype}      % microtypography
\usepackage{multirow}
\usepackage{lipsum}
\usepackage{graphicx}
\usepackage{amsmath}
\usepackage{nccmath}
\usepackage[super]{nth}
\usepackage{float}
\usepackage{amsmath}
\usepackage{tablefootnote}
\graphicspath{ {./images/} }
\usepackage{threeparttable}

\title{Spatio-Temporal Graph Convolutional Networks for Road Network Inundation Status Prediction during Urban Flooding}

\author{
 Faxi Yuan \\
  Urban Resilience.AI Lab\\
  Zachry Department of Civil and \\Environmental Engineering\\
  Texas A\&M University\\
  College Station, TX 77843 \\
  \texttt{faxi.yuan@tamu.edu} \\
  \And
 Yuanchang Xu \\
  Urban Resilience.AI Lab\\
  Department of Computer Science and Engineering\\
  Texas A\&M University\\
  College Station, TX 77843 \\
  \texttt{yuanchangxu@tamu.edu} \\
  \And
 Qingchun Li \\
  Urban Resilience.AI Lab\\
  Zachry Department of Civil and \\Environmental Engineering\\
  Texas A\&M University\\
  College Station, TX 77843 \\
  \texttt{qingchunlea@tamu.edu} \\
  \And
 Ali Mostafavi \\
  Urban Resilience.AI Lab\\
  Zachry Department of Civil and \\Environmental Engineering\\
  Texas A\&M University\\
  College Station, TX 77843 \\
  \texttt{amostafavi@civil.tamu.edu} \\
}

\begin{document}
\maketitle
\begin{abstract}
The objective of this study is to predict the near-future flooding status of road segments based on their own and adjacent road segments’ current status through the use of deep learning framework on fine-grained traffic data. Predictive flood monitoring for situational awareness of road network status plays a critical role to support crisis response activities such as evaluation of the loss of access to hospitals and shelters. Existing studies related to near-future prediction of road network flooding status at road segment level are missing. Using fine-grained traffic speed data related to road sections, this study designed and implemented three spatio-temporal graph convolutional network (STGCN) models to predict road network status during flood events at the road segment level in the context of the 2017 Hurricane Harvey in Harris County (Texas, USA). Model 1 consists of two spatio-temporal blocks considering the adjacency and distance between road segments, while Model 2 contains an additional elevation block to account for elevation difference between road segments. Model 3 includes three blocks for considering the adjacency and the product of distance and elevation difference between road segments. The analysis tested the STGCN models and evaluated their prediction performance. Our results indicated that Model 1 and Model 2 have reliable and accurate performance for predicting road network flooding status in near future (e.g., 2-4 hours) with model precision and recall values larger than 98\% and 96\%, respectively. With reliable road network status predictions in floods, the proposed model can benefit affected communities to avoid flooded roads and the emergency management agencies to implement evacuation and relief resource delivery plans.
\end{abstract}

% keywords can be removed

\section{Introduction}
Road networks act as the backbone of modern cities to enable the mobility of goods, information and people (Dong et al. 2020a; Pregnolato et al. 2017). Road network failures particularly during disasters not only drastically reduce affected communities’ access to essential services such as hospitals and shelters (Yuan et al. 2021a; Dong et al. 2020b), but also brings increasing burdens to the implementation of search and rescue strategies by emergency management agencies (Helderop and Grubesic 2019). Flood impacts on road networks further cause disruptions in industrial productions, logistics and businesses (Jenelius et al. 2006), through both direct impacts (e.g., physical damages to road segments) and indirect impacts (e.g., congestions in traffic flow) (Zhang et al. 2019; Brown and Dawson 2016). Additionally, road network failures have been identified as one of the predominant causes of deaths in urban regions during floods, as vehicles were driven into flooded road segments (Drobot et al. 2007). The ability to have foresights regarding flooding status of road network in the hours to come is critical for not only the affected communities to avoid flooded roads and reduce losses but also the emergency management agencies to understand which communities have lost access to any essential facilities such as hospitals and groceries (Yuan et al. 2021a). The capability for predicting the near-future inundation of roads is an essential part of the predictive flood monitoring for situational awareness, which is defined as the capability to have foresights about the near-future (in two to six hours) flooding status based on the predictions with current flooding status.

\subsection{General flood risk assessment} 
Existing regional flood risk assessment approaches mainly employ hydraulic and hydrologic models (Versini et al. 2010). Using hydraulic simulation tools, a number of studies have developed systems to predict road network status during floods (Chang et al. 2018). Versini et al. (2010) developed a road inundation warning system (RIWS) to assess flooding risk of the road network. Based on the RIWS proposed by Versini et al. (2010), Naulin et al. (2013) have developed a distributed hydro-meteorological forecasting approach to detect road inundation risks. Yin et al. (2016) integrated a hydrodynamic model called FloodMap-HydroInundation2D for simulating overland flow and flood inundations on road network. With simulated flood scenarios, the study proposed and employed a new algorithm and proxy to evaluate the flood impacts on road network. Wang et al. (2019) designed a failure propagation model to explore how local floods caused large-scale disruption failures in the road network, where the flood scenarios were simulated by the CaMa-Flood global river flood model. However, the hydraulic and hydrologic models in these studies are mainly useful for preparedness and hazard mitigation stages in the flood risk assessment (Versini et al. 2010), while predictive flood monitoring for situational awareness (during response stage) requires the predictive capability to enhance the understanding of the near-future (in two to six hours) flooding status of road network based on their own inundation status and the inundation status of their adjacent roads.

\subsection{Flood monitoring for situational awareness} 
Flood monitoring for situational awareness refers to the real-time or near real-time monitoring of flooding status (e.g., Yuan et al. 2021a). The existing flood monitoring systems use physical sensors to collect data regarding water elevation in channels and rivers. However, the number of physical sensors is limited and cannot provide full observability of the entire regions. In addition, the physical sensors do not provide insights regarding inundation status of roads. To bridge this gap, some studies have used crowdsourcing and social media data for monitoring of flooding status on the road network (Fan and Mostafavi 2019; Blumberg et al. 2015). Schnebele et al. (2014) employed the crowdsourced photos and volunteered geographic data to monitor road flooding status through a geostatistical interpolation method in Hurricane Sandy. Blumberg et al. (2015) utilized hundreds of photos related with Hurricane Sandy from volunteers to simulate the flood inundations in Hoboken and Jersey City (USA). Yin and Wilby (2016) used the crowdsourced data to validate their simulated flood scenarios and the associated impacts on land subsidence in Shanghai China. In addition, recent studies have used social media data to capture road flood status. Chen et al. (2020) employed Twitter data to evaluate flood impacts on highways during Hurricane Harvey. Yuan et al. (2021a) applied Twitter data to assess road network function losses in Hurricane Florence. Both of these studies have utilized the highway/road term/name lexicons to identify the related social media data. Chen et al.’s study focused on 10 highways, while floods impacts on low level roads were not investigated. Yuan et al.’ study has identified 13 flood impacted roads in Wilmington North Carolina. Most road segments’ status cannot be comprehensively monitored with social media data (e.g., Yuan et al. 2021a). These studies based on crowdsourcing and social media data, however, are mainly used for the capture of current or past (a few hours ago) flooding status of road network and cannot provide predictions of the near-future flooding status of road network (i.e., whether a certain road segment would be inundated given the its current inundation status and the inundation status of other roads).

\subsection{Point of departure} 
The objective of this research is to advance predictive flood monitoring for situational awareness of road network status through the use of spatio-temporal graphic convolutional networks (STGCN) models with high-resolution traffic data. The proposed STGCN models aim to make predictions of road segments flooding status based on their own and adjacent road segments’ current flooding status. The number of studies focusing on predictive flood monitoring of road networks are rather limited. Among the few studies available in the exiting literature, Fan et al. (2020) proposed a contagion model for flood propagation and recession in the road network of Harris County (Texas, USA). The study utilized high-resolution traffic data to infer the inundation status of road segments (road segments with no traffic speed were shown to be inundated). Hence, traffic speed data during floods can be a reliable indicator for road network status during floods and using models for traffic flow predictions can be adjusted to derive foresights about road network flooding status. 
Yu et al. (2018) have adjusted the traditional convolutional models to graph-structured data (non-Euclid domain) for predicting traffic congestions at specific road segment level with considering the road network typology (i.e., spatial road network structure). Inspired by Yu et al. (2018), this study implements the STGCN models for predicting the near-future road network flooding status. To resolve the limitations in the assumptions of undirected graph structure in Yu et al. (2018), we introduce the elevation difference as an adjacency category to capture the spatial features of the road network. The implementation of elevation difference is consistent with flood propagation mechanism in the road network, as floods would not propagate from road segments at lower elevations to their adjacent road segments at higher elevations. This consideration enables defining the direction of graph structures for our road network. In this study, three architectures are designed for the STGCN models, which are utilized for predicting the near-future flooding status of the road network at the road segment level. Using the road network of Harris County (Texas, USA) in the context of the 2017 Hurricane Harvey flooding, we validate our proposed STGCN models.
The remainder of the paper is organized as follows. Section 2 describes the methods and materials for the design of STGCN models and how to capture the spatio-temporal features of traffic condition data. Section 3 provides the road network status prediction results and the performance of the STGCN models. Section 4 discusses the contributions and limitations of research findings and concludes this paper. 

\section{Materials and Methods}
\subsection{Traffic condition prediction model} 
Road condition prediction (such as traffic prediction) is a time-series prediction problem, where we can make forecasting of the most probable traffic condition variables such as speed and traffic flow in the future S time steps with previous N traffic condition observations. The model was represented by Eq. (1).

\begin{equation}
\label{eq:1}
v_{t+1}^\prime,\ldots,v_{t+S}^\prime=\begin{matrix}arg\ max&\ log\ P\left(v_{t+1},...,v_{t+S}\middle| v_{t-N+1},...,v_t\right)\\v_{t+1},...,v_{t+S}&\\\end{matrix}
\end{equation}
Where $v_t{\in}{{\displaystyle \mathbb {R}}}^M$ is an observation vector of speed for \emph{M} road segments at time step \emph{t} and each vector element represents a historical speed for a road segment; ${v_t}^\prime$ represents the predicted speed vector with highest probability at time step \emph{t}. 

Zhang et al. (2019) has classified these methods into two categories: model-driven and data-driven. Model-driven methods depend heavily on prior knowledge of the urban environment (Abadi et al. 2015) and the corresponding simulation systems are not easily generalized to other environments (Manley et al. 2014). Data-driven approaches become more popular with the increasing availability of urban traffic data. This category generally includes two parametric and nonparametric approaches (Zhang et al. 2019). The parametric approaches were mainly applied for predicting short-term traffic conditions in small road networks (e.g., Min et al. 2009; Adeli and Jiang 2008), and most of them suffer from large computation load when implemented in large road networks. Nonparametric approaches study the relationship between the inputs and outputs to predict traffic conditions and commonly used models include artificial neural networks (Boto-Giralda et al. 2010) and support vector regression (Haworth et al. 2014). These commonly used models have limited prediction accuracy due to their shallow architectures while traffic conditions have characters including complex spatial dependence and nonlinear temporal dynamics (Lv et al. 2015).

In addition to traditional models of nonparametric approaches, deep learning models have attracted increasing attention for traffic condition predictions. Lv et al. (2015) employed a deep learning model for short-term traffic flow prediction and found their model presented higher accuracy than other standard machine learning models. However, road network’s physical characters were not captured in their deep learning model. Considering both temporal and spatial characters of traffic conditions, Zhang et al. (2016) utilized convolution neural network (CNN) to establish the deep learning model for traffic prediction. Thereafter, various studies integrated CNN with the recurrent neural network (RNN) or long-short-term memory (LSTM) for spatio-temporal traffic condition predictions (Lu et al. 2020; Polson and Sokolov 2017; Yang et al. 2016; Yu et al. 2017). As CNNs only apply to a Euclid domain, all the mentioned deep learning models performed their predictions of traffic conditions at grid level, where road networks were divided into a set of adjacent grids. Traffic conditions of specific road segments were not available. As road segment status were defined according to their traffic conditions during floods, these deep learning methods cannot be used for predictions of status of road segments (e.g., traffic condition). To resolve the limitations of the CNNs, recent studies have adjusted the traditional convolutional models to graph-structured data (non-Euclid domain) for traffic condition predictions (Defferrard et al. 2016). There are two main approaches to adjust the traditional convolutional filters to graph-structured data: extension of the convolution’s spatial definition (Niepert et al. 2016), and operation with graph Fourier transformation in spectral domain (Defferrard et al. 2016). The former applies to both directed and undirected graphs, while it has limitations in terms of the selected neighbor quantity (Zhang et al. 2019). The later suits undirected graphs and has no limitations in terms of the number of selected neighbors. Yu et al. (2018) have implemented the second approach for traffic speed predictions in multi-time steps.

\begin{figure}[h]
 \centering
 \includegraphics[width=0.35\textwidth]{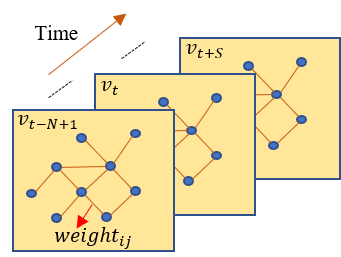} 
 \vspace{-.2in}
 \caption{Graph-structured traffic speed data (adjusted from Yu et al. (2018)).}
 \label{fig:Overview}
 \vspace{-.05in}
\end{figure}

Inspired by Yu et al. (2018), this study modeled road network with traffic condition (i.e., traffic speed) as a graph structure with concentration on structured traffic time series. In the graph network, each historical traffic speed record $v_t$ (nodes) is connected by pairwise connection, and edges with weights $W_{ij}$ represent the connection status (i.e., spatial features of road network) between road segments, which was illustrated in Figure 1. In Yu et al. (2018), $W_{ij}$ refers to an element of weight matrix, which was calculated with the adjacency matrix. This study introduces three adjacency matrix categories. The first adjacency matrix category was computed with Eq. (2), and was further used for calculating the first weight matrix for further graph convolutions (section 2.2). This adjacency matrix can only capture the 2-D distance characters between road segments, while prediction of road inundation conditions in floods needs to consider the flood character as flood water can only flow from road segments at higher-elevation places to that at lower-elevation places. Therefore, our approach introduces elevation difference to build the second adjacency matrix, which can be calculated by Eq. (3). The second adjacency matrix category is applied to computing the second weight matrix. Integrating both distance and elevation difference, we propose the third adjacency matrix as presented in Eq. (4), which creates the third weight matrix.

\begin{equation}
\label{eq:2}
  First \; Adjacency_{ij} =
  \begin{cases}
    e^{-\frac{{distance}_{ij}^2}{100}},& \text{if $i\neq j$ and $e^{-\frac{{distance}_{ij}^2}{100}}$>0.3}\\
    0, & \text{otherwise}
  \end{cases}
\end{equation}

\begin{equation}
\label{eq:3}
 {Second \; Adjacency}_{ij} = 
 \begin{cases}
    +\infty,& \text{if ${elevation}_{ij}$ < 0, and $i\neq j$}\\
    e^{-\frac{{elevation}_{ij}^2}{100}},& \text{if ${elevation}_{ij}$\ >0, and $i\neq j$}\\
    0, & \text{if $i=j$}\\
\end{cases}
\end{equation}

\begin{equation}
\label{eq:4}
{Third \;Adjacency}_{ij} = {First \; Adjacency}_{ij}\ \times {Second \; Adjacency}_{ij}
\end{equation}

Accordingly, the road graph can be represented by {$\mathcal{G}_t$}=({$\mathcal{V}_t$},\ $\mathcal{E}$,W), where $\mathcal{V}_t$ refers to the historical traffic speed of \emph{N} road segments at time step t and is represented by a set of nodes, $\mathcal{E}$ reflects the connection status between road segments and a set of edges, and W can be calculated with the three adjacency matrix categories as defined in Eq. (2-4). As a result, each historical traffic speed record $v_t$ refers to a graph signal of road graph $\mathcal{G}_t$. With the defined road graph, we implemented the convolutional neural network (CNN) and deigned the models with the spatio-temporal graphic convolutional networks to predict road traffic conditions, and subsequently road status in floods, which are further discussed in the following sections. 

\subsection{Description of model architectures} 
We designed three models with the spatio-temporal graphic convolutional networks (STGCN) modified from the original STGCN model proposed by Yu et al. (2018). The network architectures for these three models are illustrated in Figure 2. 

\begin{figure}[hbt!]
\centering
\includegraphics[width=\linewidth]{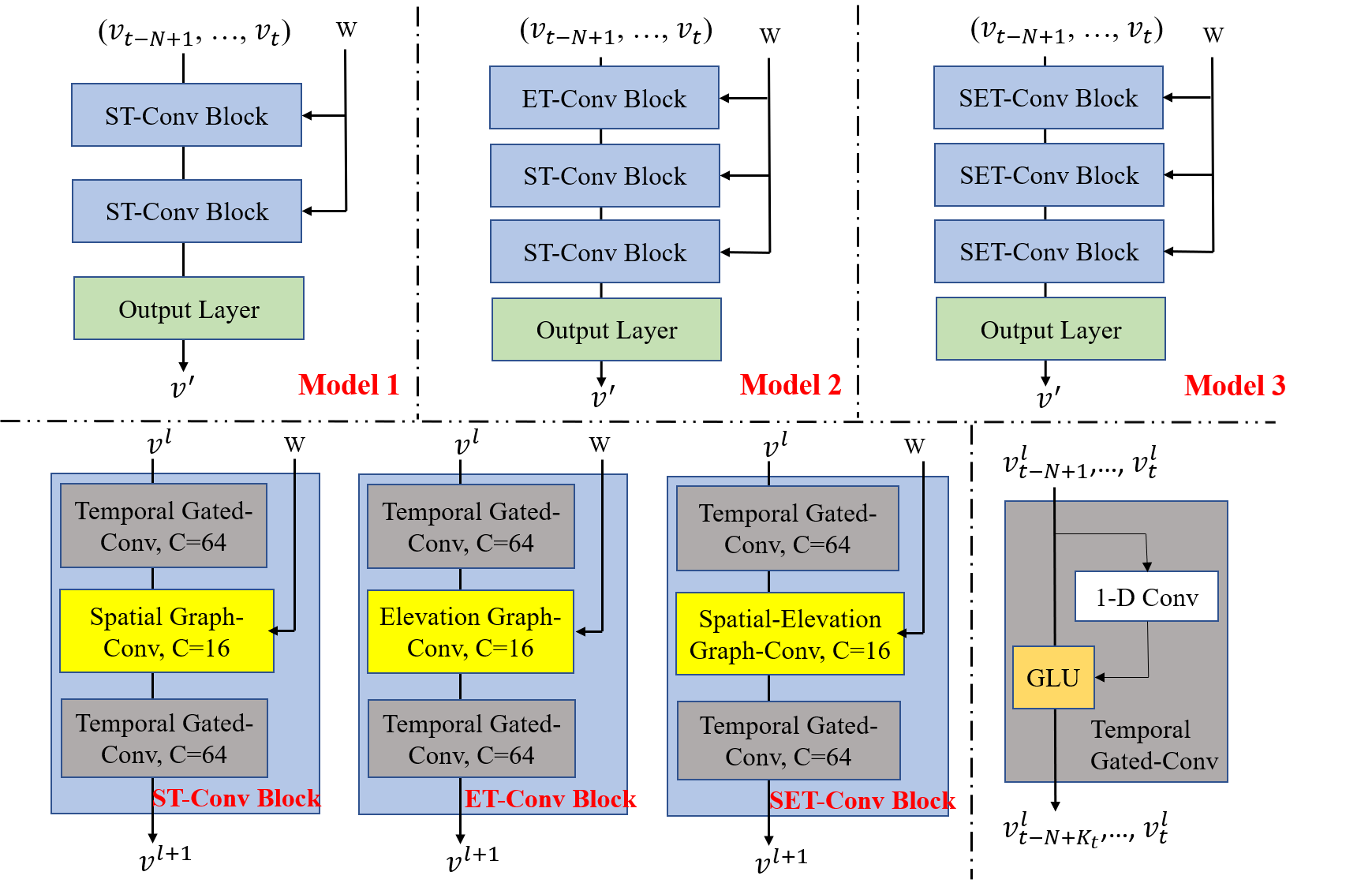}
\caption{Architectures of spatio-temporal graph convolutional network models}
\end{figure}

Model 1 consists of two spatio-temporal blocks (ST-Conv block) and a fully connected output layer in the end. Each ST-Conv block consists of two temporal gated convolution layers at the front and end, and one spatial graph convolution layer built with the first adjacency matrix category (Eq. (2)) in the middle. With Model 1, two ST-Conv blocks process the road traffic data input $v_{t-N+1}$, …, $v_t$ (i.e., vehicle speed) to explore the spatial and temporal dependencies. The output layer captures the comprehensive features from the processed information to predict the future road traffic condition $v^\prime$ and inundation status.

Model 2 includes an elevation-temporal blocks (ET-Conv block), two spatio-temporal blocks (ST-Conv block), and a fully connected output layer in the end. The ST-Conv block possesses the same layers as those in the Model 1. Each ET-Conv block contains two temporal gated convolution layers at the front and end, and one elevation graph convolution layer built with the second adjacency matrix category (Eq. (3)) in the middle. Each ET-Conv block processes the input feature data as ST-Conv block does in Model 1, and so does the output layer.

Model 3 employs three spatio-elevation-temporal blocks (SET-Conv block) and an output layer. Each SET-Conv block consists of two temporal gated convolution layers at the front and end, and one spatial-elevation graph convolution layer built with the third adjacency matrix category (Eq. (4)) in the middle. Similar to the EST-Conv block in Model 1, each SET-Conv block processes the input data to investigate the spatial-elevation and temporal dependencies, and the output layer makes the prediction of future road traffic conditions $v^\prime$ and inundation status.

\subsection{Graph CNNs for extracting spatial and elevation features} 
With the defined road graphs, this study employed the graph CNNs to extract the spatial patterns and features. As defined by Yu et al. (2018), this research employed $\ast_\mathcal{G}$ to represent the graph convolution operator to multiply a signal $x \in \mathbb{R}^M$ with a kernel $\Theta$ as presented in Eq. (5). Through the involvement of three W categories computed with the three adjacency matrix categories as defined in Eq. (2-4), spatial patterns and features including distance, elevation difference and the product of distance and elevation difference between road segments were captured by the graph CNNs.

\begin{equation}
\label{eq:5}
\Theta\ \ast_\mathcal{G}{x}\ =\ \Theta(L)x=\Theta(U\mathrm{\Lambda}U^T)x=U\Theta(\mathrm{\Lambda})U^{T}x
\end{equation}

Where $L=I_M-D^{-\frac{1}{2}}WD^{-\frac{1}{2}}=U{\mathrm{\Lambda}}{U^T}\in\mathbb{R}^{M\times M}$ represents the normalized graph Laplacian matrix; $I_M$ is an identity matrix; $D$ was the diagonal degree matrix with ${D_{ii}}={\sum_{j}} {W_{ij}}; U{\in}{\mathbb{R}^{M\times M}}$ is the Fourier basis and referred to the matrix of eigenvectors for the normalized graph Laplacian L; $\mathrm{\Lambda}{\in}\mathbb{R}^{M\times M}$ is denoted as the diagonal matrix of eigenvalues of the normalized graph Laplacian L; $\Theta(\mathrm{\Lambda})$ also represents a diagonal matrix.

To reduce the computation load of kernel $\Theta$ with graph Fourier transform, this step utilizes the Chebyshev polynomials approximation method (Hammond et al. 2011). Accordingly, the graph convolution in Eq. (5) can be adjusted as Eq. (6), where the computation cost can be reduced (Defferrard et al. 2016).
\begin{equation}
\label{eq:6}
\Theta\ \ast_\mathcal{G}x =\Theta(L)x\approx\sum_{k=0}^{K-1}{\Theta_kT_k(\widetilde{L})x}
\end{equation}

Where $\widetilde{L}=\ 2L/\lambda_{max}-{I_N}$ was the scaled Laplacian matrix of $\emph{L}$; $T_k(\widetilde{L}) \in \mathbb {R}^{M\times M}$ referred to the Chebyshev polynomial of order k evaluated at $\widetilde{L}$; $\emph{K}$ denotes the kernel size of graph convolution; $\theta \in \mathbb{R}^K$ refers to the polynomial coefficient vector.

With the approximation of kernel $\Theta$, we generate the graph convolutions to capture the spatial and elevation features. Our defined graph convolution operator $\ast_\mathcal{G}$ was based on 1-D signal $x \in \mathbb{R}^M$, while it can also be applied to multi-dimensional tensors. For a signal within $C_{in}$ channel, we have a signal matrix $X \in \mathbb{R}^{M\times C_{in}}$ and the 2D graph convolution was generalized by Eq. (7), where $\mathcal{Y}_j \in \mathbb{R}^M$.
\begin{equation}
\label{eq:7}
\mathcal{Y}_j=\sum_{i=1}^{C_{in}}{\Theta_{i,j}\left(L\right)x_i}, 1\le j\le C_{out}
\end{equation}

Where $C_{in}$ and $C_{out}$ are the input and output sizes respectively for their corresponding feature maps; the Chebyshev coefficients $\Theta_{i,j} \in \mathbb{R}^K$ applied to all the $C_{in} \times C_{out}$ vectors. Accordingly, the graph convolutions for 2-D variables are illustrated by $\Theta\ {\ast_\mathcal{G}X}$, where $\Theta \in \mathbb{R}^{K\times C_{in}\times C_{out}}$. In the traffic prediction and road inundation model, the input of traffic speed contain N time steps and each time step $\emph{t}$ has historical traffic speed $v_t$. Each $v_t$ is treated as a $M\times C_{in}$ matrix, where $\emph{M}$ refers to the number of road segments while $C_{in}$ denotes the number of road condition features. In our case, we only consider traffic speed as road condition feature so that $C_{in}=1$. As we had N time steps, our input signal $\mathcal{X} \in \mathbb{R}^{N\times M\times C_{in}}$ was a 3-D feature, and the updated graph convolution was generalized in Eq. (8).

\begin{equation}
\label{eq:8}
\Theta \ast_\mathcal{G}\mathcal{X} = \Theta(L)\mathcal{X} 
\end{equation}

As Figure 2 (lower right part) shows, there is a 1-D causal convolution (kernel width as $K_t$) and a gated linear units (GLU) as a non-linearity in the temporal layer. For the gated CNNs to capture the temporal features, we have used the same method as described in Yu et al. (2018), which were not introduced here.

\subsection{Description of convolutional blocks} 
To integrate both spatial and temporal features, we established the elevation-temporal convolutional block (ET-Conv block), spatio-temporal convolutional block (ST-Conv block), elevation-spatio-temporal blocks (EST-Conv block), and spatio-elevation-temporal blocks (SET-Conv block). The structures of these blocks are illustrated in Figure 2 (left of lower). Specifically, in the ET-Conv block, the elevation layer is in the middle of two temporal layers to realize the quick propagation of elevation state by temporal convolutions within the graph convolution. Within each ET-Conv block, we implemented the layer normalization to prevent overfitting issue. Accordingly, the structures of ST-Conv and SET-Conv blocks are similar as those of the ET-Conv block, where quick propagations of spatial and spatial-elevation states are achieved respectively, and the layer normalization strategy is applied to both of them. Compared with the structure of the SET-Conv blocks, the EST-Conv block has four layers, and the spatial and elevation layers are implemented separately. Similarly, the elevation and spatial layers are in the middle to connect the two temporal layers, where rapid propagations of elevation and spatial states are achieved, and the layer normalization is utilized.

As mentioned in section 2.3, both input and output of these four convolution blocks are 3-D tensors. With the input $v^l \in \mathbb{R}^{N\times M\times C^l}$ ($C^l$: channel \emph{l}) of block l, we computed the output $v^{l+1} \in \mathbb{R}^{(N-2(K_t-1))\times M\times C^{l+1}}$ with Eq. (9).

\begin{equation}
\label{eq:9}
v^{l+1}= \Gamma_{low}^l{\ast}_\mathcal{T} ReLU(\Theta^l{\ \ast}_\mathcal{G}(\Gamma_{up}^l{\ \ast}_\mathcal{T}v^l)) 
\end{equation}
Where ${\Gamma_{up}}^l$ and ${\Gamma_{low}}^l$ refer to the upper and lower kernel, respectively, in the block $\emph{l}$; ${\ast}_\mathcal{T}$ denotes gated convolution operator; $\Theta^l$ is the graph convolution kernel in the block $\emph{l}$; $ReLU \left(\bullet\right)$ represents the rectified linear units function (Li and Yuan 2017).

Across the three models (upper of Figure 2), an additional temporal layer is assigned after two EST-Conv blocks in Model 1, so are the assignment after one ET-Conv block and two ST-Conv blocks in Model 2, and the attachment after three SET-Conv blocks in Model 3. This temporal layer projects the outputs of the last EST-Conv block in Model 1, and that of the last ST-Conv block in Model 2 and the last SET-Conv block in Model 3. Thereafter, we utilize the final outputs $FO\in \mathbb{R}^{M\times c}$ from these three models, respectively, with a linear transformation across the c channels to make predictions of traffic speed for the M road segments (Eq. (10)).
\begin{equation}
\label{eq:10}
v^\prime\ =FOw\ +\ b 
\end{equation}
Where $w \in \mathbb{R}^c$ referred to a weight vector and b was the bias. The predicted speed will be used as proxy for the inundation status of roads as we will explain the following sections.

\subsection{Model evaluation}
We implemented a two-step evaluation for the proposed STGCN models. For the model evaluations, we introduced a binary value to assign the road segment statuses. Fan et al. (2020) have found that the null average speed appeared only in the flood period (i.e., Hurricane Harvey) within the IRINX traffic dataset. In this study, we denoted the road segments with null average speed as flood inundated roads in our training dataset. However, the implementation of the STGCN models requires the features from the training dataset (i.e., average speed) without null values (Yu et al. 2018). To resolve this issue, we utilized the feature scaling function (sklearn library in Python) to normalize our training dataset. Accordingly, there is no null average speed from the predictions of the STGCN models so that we cannot use null average speed to validate the prediction results (i.e., road segment flooding inundations). Hence, we defined a traffic speed threshold through the comparison of the average speed of road segments and their corresponding historical average speed. If the ratio of observed/predicted traffic speed and historical traffic speed is below such threshold, we denoted the observed/predicted road segment status as flooded (Pregnolato et al. 2017). Otherwise, road segment status is assigned as not flooded (i.e., 0). In the first step, we compared the predicted road segments status with their corresponding observed status and employed two metrics including the mean absolute error (MAE) and root mean square error (RMSE) to evaluate the performances of these models. The computations of the MAE and RMSE until time step t were conducted using Eq. (11-12). Since most road segments in Harris County were not flooded in Hurricane Harvey (Figure 4), MAE and RMSE may not fully reflect the model performances for flooded road status predictions. Therefore, we implemented the second-step model performance evaluation with concentration on the predictions of flooded road segments. In the second step, we employed the precision and recall to evaluate the performances of our three models for identifying the flooded road segments by Eq. (13-14).
\begin{equation}
\label{eq:11}
{MAE}_t=\frac{\sum_{i=1}^{M}\left|{rs}_{t-i}^\prime-{rs}_{t-i}\right|}{S}
\end{equation}

\begin{equation}
\label{eq:12}
{RMAE}_t=\sqrt{\sum_{i=1}^{M}\frac{{({rs}_{t-i}^\prime-{rs}_{t-i})}^2}{S}}
\end{equation}

Where ${rs}_{t-i}^\prime$ referred to the predicted status for road segment \emph{i} at time step \emph{t}; ${rs}_{t-i}$ was the observed status for road segment \emph{i} at time step \emph{t}; \emph{S} was the number of time steps.

\begin{equation}
\label{eq:13}
{Precision}_t=\ \frac{true\ positive}{true\ positive\ +\ false\ positive}
\end{equation}

\begin{equation}
\label{eq:14}
{Recall}_t=\ \frac{true\ positive}{true\ positive\ +\ false\ negative}
\end{equation}
Where \emph{true positive} denotes the situation where the models correctly predicted the flooded segments; \emph{false positive} refers to the outcome where the models incorrectly predicted the flooded road segments, while \emph{false negative} is for instanced that where the models incorrectly predicted the not flooded road segments.

\section{Results}
\subsection{Data description of study context}
To illustrate the implementation and performance of the proposed graph convolutional networks for road status prediction in floods, we employed high resolution data related to traffic conditions of the roads in Harris County, Texas, during Hurricane Harvey in 2017. Hurricane Harvey made landfall in Harris County on August 26 and caused severe floods. The floods continued from August 27, 2017 to September 4, 2017. Accordingly, we defined our study timeline from August 26, 2017 to September 4, 2017, where we collected the traffic condition data for 19,712 road segments in Harris County from the private company INRIX which collects location-based data from both sensors and vehicle. The INRIX traffic data contains the average traffic speed of each road segment at 5-minute interval and their corresponding historical average traffic speed. The traffic data from INRIX covers all available road roads—from interstates to intersections, and from country roads to neighborhoods. Each road segment’s identification information such as name, geographic locations defined its head and end coordinates, and length, is also available from the INRIX data set. In this study, we use the traffic data from August 27, 2017 to September 1, 2017 as our training dataset, and that at 2:00 am, 4:00 am, and 6:00 am on September 2, 2017 as our testing dataset.

Flooded road segments due to Hurricane Harvey can be identified by detecting the road segments with NULL values for their average traffic speed. As the ATGCN models cannot accept null average speed as inputs, we normalized the null average speed for our training dataset with feature scaling function. For the predictions from the STGCN models, we calculated the ratio between predicted traffic average speed and their historical average speed with  $\frac{average \; speed}{historical \;average \; speed}\times100\%$, and set the threshold as $10\%$ to denote road segments experiencing very low speed. If the ratio is lower than $10\%$, the road segments could be flooded (still possible). The commonly allowed daytime vehicle speed is 70 miles/hour (i.e., 112.7 km/hour) in Harris County, Texas (City of Houston 2021). Accordingly, maximum average speed from our defined threshold is 11.3 km/hour (i.e., $10\% \times$ 112.7 km/hour). Using the vehicle speed $(\upsilon(w))$-flood water depth (w) function $(\upsilon(w) = 0.009w^2-0.5529w+86.9948)$ in Pregnolato et al. (2017), we can compute the water depth for 11.3 km/hour is 30.7 cm, which can be sufficient to stall a commonly used cars (excluding high water trucks). From the vehicle speed-flood water depth function, water depth increases as the vehicle speed increases (Pregnolato et al. 2017), which indicates using $10\%$ as a threshold to denote road flooding inundation status is reasonable.

Considering the computation costs when using the graph convolutional networks, we divided the road network of Harris County into five clusters according to the distribution of channels to the watersheds (Dong et al. 2020c). The distribution of the five clusters for the road segments was illustrated in Figure 3. We trained our proposed models with traffic condition data from August 26, 2017 to September 1, 2017 (i.e., seven days) in each road cluster, and evaluated their performances by aggregating all the predicted results in the whole county.

\begin{figure}[hbt!]
\centering
\includegraphics[width=0.75\linewidth]{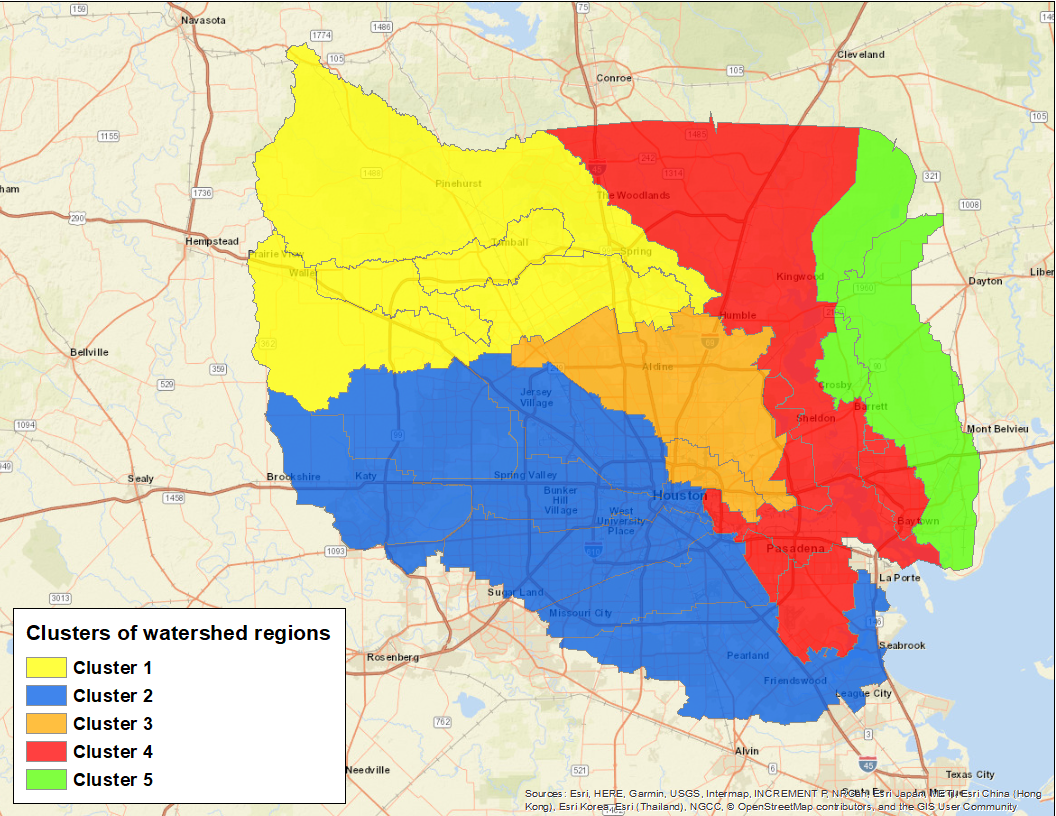}
\caption{Distribution of five clusters for the road segments in Harris County}
\end{figure}

\subsection{Data description of study context}

\subsection{Prediction results}
With our defined traffic speed threshold, this study denoted the road status with our predicted and observed traffic speed respectively. As indicated in section 2.5, we first used the mean absolute error (MAE) and root mean square error (RMSE) to evaluate the performances of our proposed STGCN models for the prediction of road segment statuses including flooded and not flooded. Considering the timeline of our training dataset is from 00:00:00 am on August 27, 2017 to 11:59:59 pm on September 1, 2017, this study made the predictions of status for each road segment at 2:00 am, 4:00 am, and 6:00 am on September 2, 2017 (i.e., two to six hours after 11:59:59 pm on September 1, 2017). Using the testing dataset at 2:00 am, 4:00 am, and 6:00 am on September 2, 2017, we computed the corresponding MAE and RMSE. The MAE and RMSE for these predictions with the three models are presented in Table 1. 

% \usepackage{multirow}

% \usepackage{tablefootnote}
% \usepackage{multirow}
% \usepackage{colortbl}

\begin{table}
\centering
\caption{Performance comparison of three STGCN models for traffic speed}
\begin{threeparttable}
\begin{tabular}{lllllll} 
\hline
\multirow{2}{*}{STGCN models} & \multicolumn{3}{l}{MSE}     & \multicolumn{3}{l}{RMSE}                                                                                                                                                               \\ 
\cline{2-7}
                              & 2:00 am & 4:00 am & 6:00 am & 2:00 am & 4:00 am                                                                                                                                                           & 6:00 am  \\ 
\hline
Model 1                       & 0.00238 & 0.00236 & 0.00409 & 0.04881 & 0.04853                                                                                                                                                           & 0.06397  \\
Model 2                       & 0.00211 & 0.00220 & 0.00277 & 0.04598 & 0.04690                                                                                                                                                           & 0.05263  \\
Model 3                       & 0.00272 & 0.00287 & 0.00344 & 0.05220 & 0.05358                                                                                                                                                           & 0.05866  \\
Fan et al. (2020)             & -       & -       & -       & -       & 0.020\tnote{*} & -        \\
\hline
\end{tabular}
\begin{tablenotes}
    \footnotesize
    \item[*] This value was based on the grid analysis of road network. Each gird refers to a square of 400-meter length and includes many road segments.  
\end{tablenotes}
\end{threeparttable}
\end{table}

The RMSE of the three STGCN models for the road status predictions at 4:00 am (i.e., 4-hour prediction) is larger than that in Fan et al. (2020). However, their predictions of road statuses were at grid level. Each road grid is a square of 400-meter length and contained many road segments. When one road segment is flooded, the grid with this road segment is denoted as flooded. However, our predictions are for individual road segments and the prediction accuracy is better than that in Fan et al. (2020). Specifically, Table 1 shows there were no significant variances in both MAE and RMSE from 2:00 am to 4:00 am for all the three STGCN models, while we can see the notable increases in these two metrics from 4:00 am to 6:00 am, particularly for Model 1 and Model 3. In addition, the values of both MAE and RMSE for Model 2 were less than their corresponding values for Model 2 and Model 3, which indicates that Model 2 had higher prediction accuracy. Hence, Model 2 performed more stable than the other two models for the general road segment statuses predictions across the three time points on September 2, 2017. As most road segments were not flooded during Hurricane Harvey in Harris County, MAE and RMSE may not accurately reflect the performances of the STGCN models for identifying the flooded road segments. Model performance evaluations with focusing on the prediction of flooded road segments were conducted in step 2.

In step 2, we focused on the predictions of flooded road segments and calculated the precision and recall for our proposed STGCN models. The results were illustrated in Table 2. According to Table 2, for the predictions of road status at 2:00 am and 4:00 am, Model 2 has demonstrated higher values for precision and recall, which means Model 2 outperformed Model 1 and Model 3. This result was consistent with that for traffic speed predictions in Table 1. Specifically, for the precision of Model 2 at 2:00 am and 4:00 am, the precision values equaled to 1.000, which meant there was no incorrect prediction of the flooded road segments and all the predictions of flooded road segments were correct. The recall values at 2:00 am, 4:00 am, and 6:00 am are not equal to 1.0. This result demonstrates that there are some flooded road segments which are not correctly captured by the Model 2.  

% \usepackage{multirow}
\begin{table}[]
\caption{Performance comparison of three STGCN models for road status prediction}
\centering
\begin{tabular}{lllllll} 
\hline
\multirow{2}{*}{STGCN models} & \multicolumn{3}{l}{Precision} & \multicolumn{3}{l}{Recall}   \\ 
\cline{2-7}
                              & 2:00 am & 4:00 am & 6:00 am   & 2:00 am & 4:00 am & 6:00 am  \\ 
\hline
Model 1                       & 1.000   & 1.000   & 0.998     & 0.987   & 0.987   & 0.986    \\
Model 2                       & 1.000   & 1.000   & 0.998     & 0.991   & 0.989   & 0.982    \\
Model 3                       & 1.000   & 1.000   & 0.997     & 0.929   & 0.923   & 0.919    \\
\hline
\end{tabular}
\end{table}
Figure 4 illustrated the true positive, false negative and false negative for the road segments from the predictions by Model 2. The light blue nodes represented the normal road segments. As the precision values at 2:00 am, 4:00 am, and 6:00 am were equal or almost equal to 1.00, there are almost no road segments with incorrect predictions as flooded. Accordingly, there is no road segment with false negative result (i.e., red nodes) in Figure 4. Also, road segments with true positive results (green nodes in Figure 4) reflected the correctly predicted flooded road segments. The recall values ranged from 0.98 to 0.99, which indicated that around $2\%$ of flooded road segments are predicted as not flooded by Model 2 (yellow nodes in Figure 4).

\begin{figure}[htbp]
    \centering
    \begin{minipage}[c]{0.75\textwidth}
        \centering
        \includegraphics[width=1\textwidth]{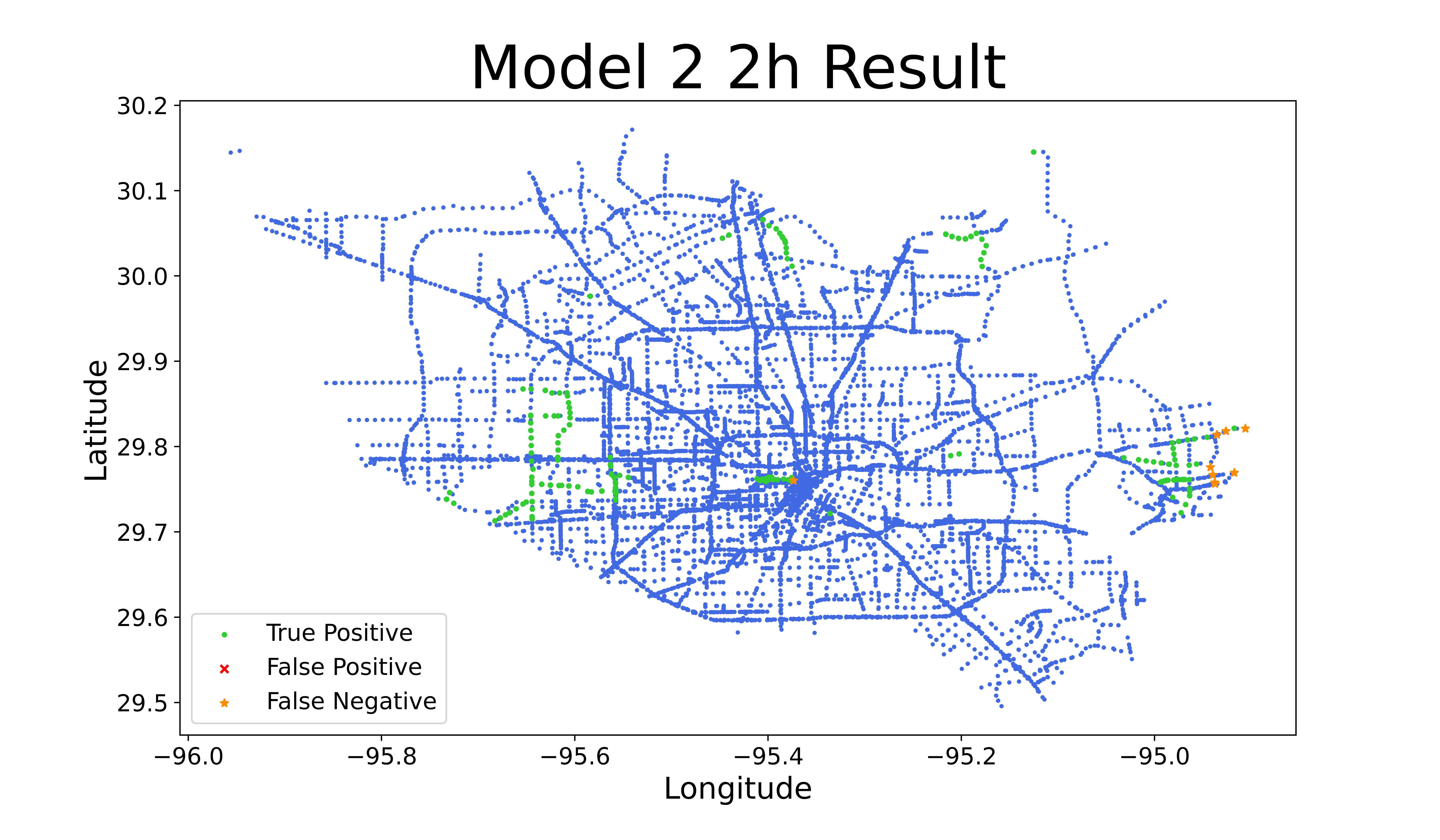}
    \end{minipage}
    
    \begin{minipage}[c]{0.75\textwidth}
        \centering
        \includegraphics[width=1\textwidth]{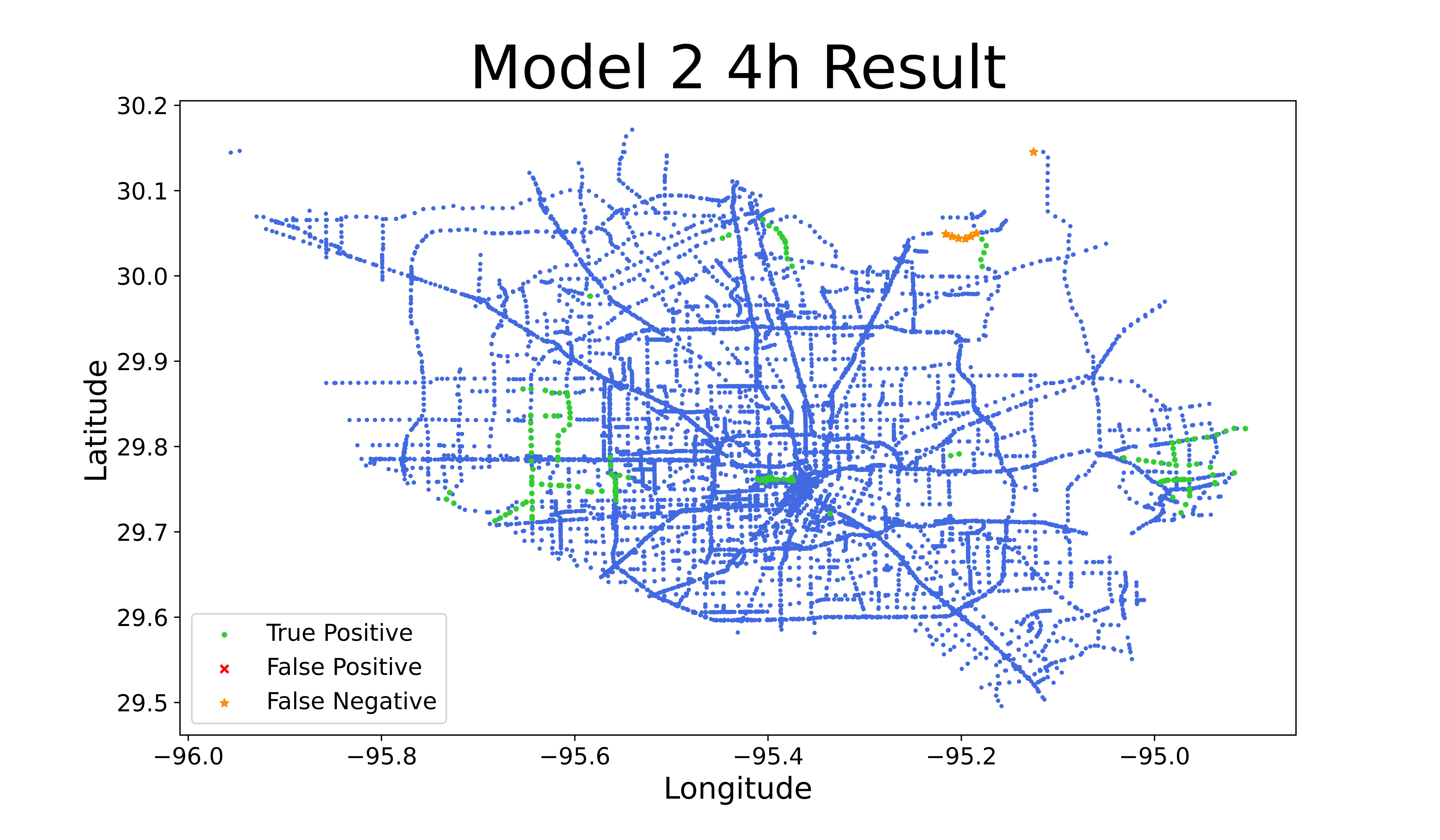}
    \end{minipage}
    
    \begin{minipage}[c]{0.75\textwidth}
        \centering
        \includegraphics[width=1\textwidth]{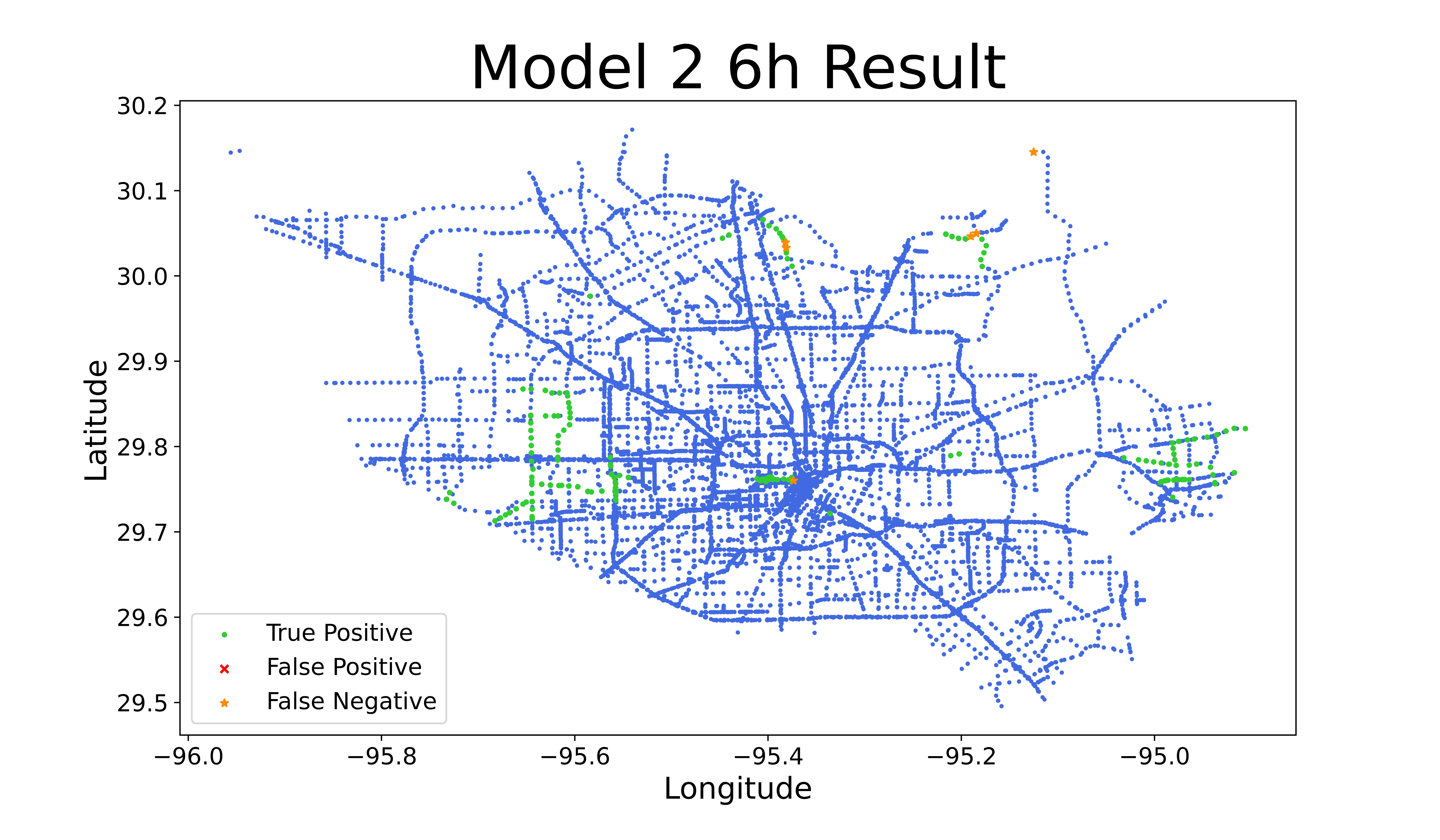}
    \end{minipage}
    
    \caption{Examples of road status predictions at 2:00 am (4a), 4:00 am (4b), and 6:00 am (4c) with the STGCN Model 2 (node represents a road segment).}
    \label{fig:foobar}
\end{figure}

In Figure 4, we can see that road segments with false negative (yellow nodes) are mainly distributed to the east of Harris County (cluster 5 in Figure 3) at 2:00 am, and that at 4:00 am and 6:00 am are mainly located in the northeast (cluster 4 in Figure 3) and north (cluster 1 in Figure 3), respectively. In addition, from 4:00 am to 6:00 am, the number of road segments with true positive results (green nodes) has decreased in the east of Harris County, which could indicate that flood recession happening in that area. Thus, with the near-future road status predictions obtained the STGCN models, emergency management agencies and crisis response managers could anticipate which areas could lose access and inform residents to avoid roads with potential inundation.

\subsection{Model adjustment}
To examine the performance of the STGCN models in terms of road network flood status predictions, this section predicts the road status for the remaining three days (i.e., 72 hours) from September 2, 2017 to September 4, 2017. The predicted road status is compared with the observed status from the IRINX dataset to compute the precision and recall at 4-hour intervals for the three STGCN models. The results are presented in Figure 5.

\begin{figure}[htbp]
    \centering
    \begin{minipage}[c]{0.49\textwidth}
        \centering
        \includegraphics[width=1\textwidth]{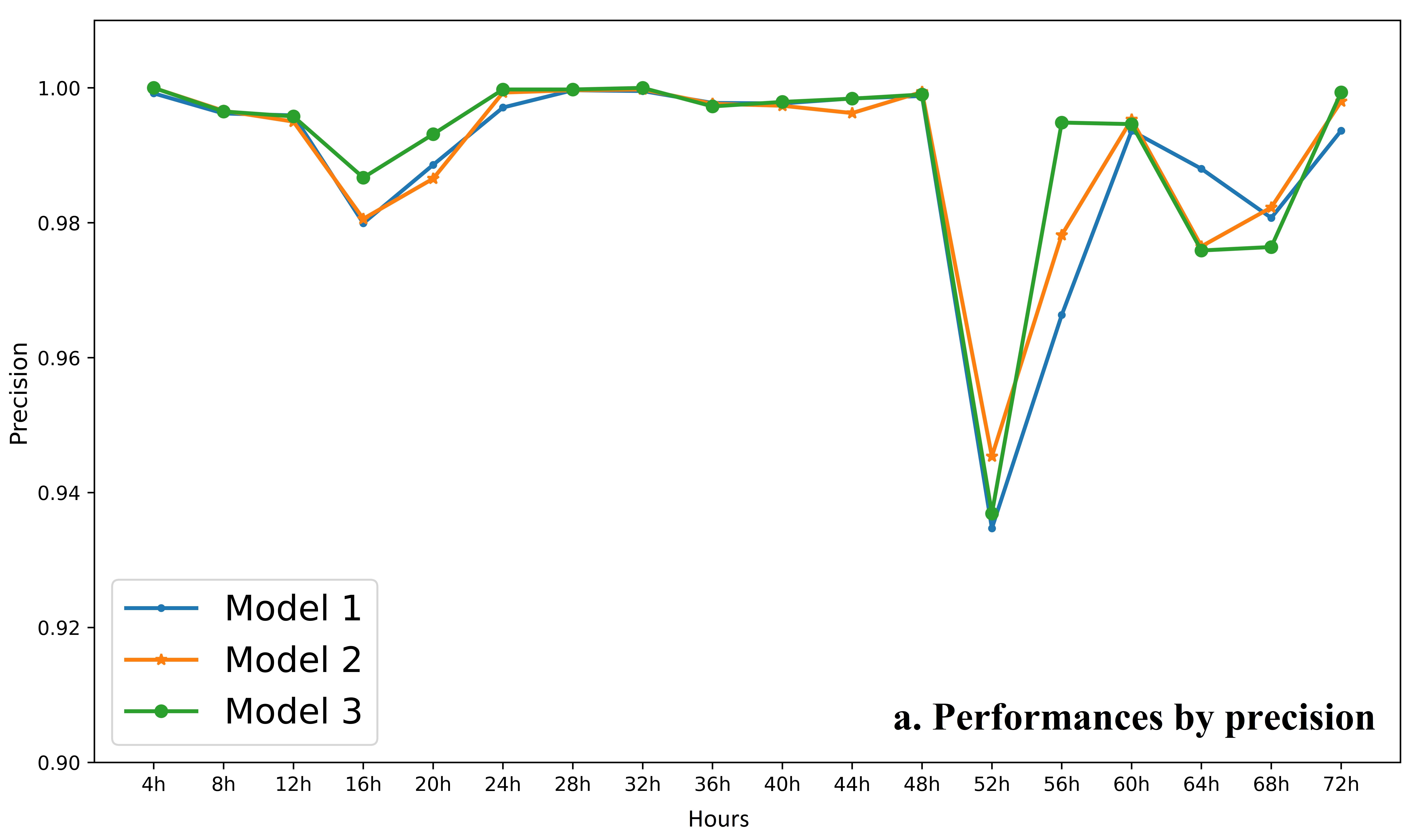}
    \end{minipage}
    \begin{minipage}[c]{0.49\textwidth}
        \centering
        \includegraphics[width=1\textwidth]{precision.jpg}
    \end{minipage}
    \caption{Prediction performances of the three STGCN models for road status in floods at 4-hour intervals. 5a: precision; 5b: recall. Hour 4h is 4:00 am on September 2, 2017, while hour 72h is 0:00 am on Sep 5, 2017.}
    \label{fig:foobar}
\end{figure}

In the left part of Figure 5, precision curves for the three STGCN models have shown the similar trends. Variances in precisions for these models presented relative stability in the first 48 hours, while a significant decrease appeared at 52-hour point.  This is reasonable as the first 48-hour traffic data (i.e., since 11:59:59 pm on September 1, 2017) were more dependent on our training dataset than the traffic data at the 52-hour point. The varying dependence reflects the temporal features of the traffic data so that the closer to the timeline of training dataset (i.e., 11:59:59 pm on September 1, 2017), the future traffic condictiones will be more dependent on the training dataset. This phenomenon indicated that all these three STGCN models have demonstrated certain stabilities in terms of precisions for road segment statuses predictions in the 48 hours. Also, the precision values for these models are very close in the 48 hours. However, precisions of the STGCN models are not stable after 48 hours and their values have seen significant variances across the three models. Hence, the results demonstrate that the three STGCN models were reliable tools for correctly identifying flooded road segments with higher precisions (i.e., >98\%) in 48 hours, while their performances were not stable after 48 hours.

In the right part of Figure 5, recall curves also show similar trend for the three STGCN models. For Model 1 and Model 2, their recall curves indicate relative stability across the 72 hours and show only one slight decrease at 52-hour point. The recall curve of Model 3 only presents stability from 16-hour point to 48-hour point and shows a notable decrease at 56-hour point. Also, we can clearly see the better performance of Model 1 and Model 2 than Model 3 in terms of their recall values. This situation reveals that Model 1 and Model 2 had less incorrect predictions of not flooded road segments than Model 3. Thus, we conclude that Model 1 and Model 2 had better performance than Model 3 as they had better stability and less incorrect predictions of flooded road segments as not flooded.

Juxtaposing both figures in Figure 5, we find that the selection of proper STGCN models for identifying flooded road segments needed to consider both model stability and their precisions and recalls. The performance of the models has shown sharp drops of stability in both precision and recall at 52-hour step, and significant variances in precisions have appeared after 48-hour time step. Considering the temporal features of traffic data, this situation can be expected as the traffic data at 52-hour point is less dependent on the training dataset than the traffic data in the 48 hours from 11:59:59 pm on September 1, 2017. The higher precisions and recalls ensure less incorrect predictions of flooded road segments and not flooded road segments, respectively. With similar stability of precisions (i.e., three STGCN models in 48 hours), we needed to select the STGCN models with higher recalls (i.e., Model 1 and Model 3). 

\section{Discussions and Conclusion}
Predictive flood monitoring for situational awareness of road network can help the affected communities to avoid flooded roads and the emergency management agencies to identify the communities losing access to any essential facilities. This study adjusted and tested the STGCN models to predict road network flooding status at road segment level based on their own and adjacent road segments’ current flooding status. The inundation status of roads is inferred from high resolution traffic data. This study considered with flood propagation characters (i.e., flood water can only propagate from high-elevation places to low-elevation places) into the STGCN models, where graph and gated CNNs were applied for capturing spatial features of road network (distance and elevation differences among road segments) and temporal features of historical traffic data, respectively. The three STGCN models were tested on the high-resolution traffic data during Hurricane Harvey in Harris County. The results show the capability of the models for the prediction of near-future road network flooding status.

This study employs variation in traffic speed as an indicator of flood status on road segments and provides a reliable approach for predicting the near-future (two to six hours to come) road network flooding status at road segment level. Compared with a very recent study which employed traffic speed to predict flood propagations in urban road network at grid level (Fan et al. 2020), the STGCN models can make predictions for specific road segments’ flooding status with considering both spatial features of road network and temporal features of traffic data. Among the three STGCN models, we can see Model 1 and Model 2 have reliable performance for predicting road network status in the future 48 hours (00:00:00 am of September 2, 2017, to 11:59:59 pm of September 3, 2017) during Hurricane Harvey. Their precisions and recalls are larger than 98\% and 96\% in the first 48 hours as we used the binary variables to define road inundation status based on our threshold. The results indicate the STGCN models can provide reliable predictions of road network flooding status in 48 hours.

For the implementation of practice, this research resolves the data limitation issue for near-future predictions of road flood inundation status. We introduced the high-resolution traffic data as proximity of road flood inundation status as this dataset could become available in a short time (e.g., one to two hours) so that we can use them to predict near-future (i.e., in two to six hours) road flood inundation status. The affected communities can use the dataset or refer to the predicted road network flooding status in the future two to six hours to select their routes to places or other essential services (e.g., hospitals and shelters) by avoiding the roads with high probability of being inundated by the flood. As existing studies have shown that driving vehicles through flood inundated roads is among the first cause of deaths in urban floods (Jonkman and Kelman 2005; Fitzgerald et al. 2010; Drobot et al. 2007), our results can help affected communities to avoid driving into flood inundated roads and further reduce deaths. On the other side, the emergency management agencies can implement such tools for the design of crisis response strategies (e.g., search and rescue activities). For instance, the emergency management agencies may have limited high water vehicles and boats to perform relief resource distribution and rescue so that they need to know which areas they can use these facilities and which areas they can use regular vehicles. Also, our results can benefit the emergency management agencies with enhanced situational awareness of communities losing accessibility to essential facilities such as hospitals. For the communities losing accessibility to hospitals, the emergency management agencies can use census data to check if there are mainly elder living in that community and further provide medical resource and support to them.

This study contributes to the body knowledge related to smart flood resilience. The increasing availability of big data and advanced developments in artificial intelligence have called up the concept of smart flood resilience (Jongman 2018). Smart flood resilience can be described as using data, models, and artificial intelligence approaches to help people better respond and react to floods through enhanced predictive flood exposure and risk mapping before floods (Dong et al. 2020b, 2020c), automated rapid impact assessment during floods (Yuan et al. 2021b; Yuan and Liu 2020, 2019, 2018a, 2018b), infrastructure failure prediction and monitoring before and during floods (Dong et al. 2020a; Fan et al. 2020), and smart situational awareness in response and recovery during and after floods (Yuan et al. 2021c, 2020a, 2020b; Podesta et al. 2020; Zhai et al. 2020; Zhang et al. 2020; Yuan and Liu 2018c). Therefore, our research contributes to the infrastructure failure prediction and monitoring during floods with concentrating on road network for augmenting smart flood resilience.

A limitation in this study is the heavy computation cost when using STGCN models to train our dataset and further make predictions of road network status during floods. Although Defferrard et al. (2016) have introduced the Chebyshev polynomials approximation to reduce the computation complexity, future efforts to reduce the computation load for applying our proposed STGCN models are still in need. Based on our current study concentrating on predicting the near-future flooding status of road network, future research can consider the predictions of congestions within the road network during floods. Lastly, the definitions of traffic speed threshold to denote road segment status during floods may vary across regions and flood events. This study used 10\% as the threshold after a detailed review of historical traffic data and reported flooded road segments in Harris County during Hurricane Harvey, while this threshold may not be generally applied to other flood events and other study regions. Future implementations of the STGCN models for road network status predictions in other regions and flood events should pay attention to the sensitivity of model outputs to this threshold value. 
Despite the limitation, this study provides a new model to predict near-future road network flooding status using high-resolution traffic data from which inundation status of roads could be inferred. This modeling approach can be generalized to other flood events and regions with proper traffic datasets for model training. The model and its predictions can support the affected communities and emergency management agencies’ response activities.

%\bibliographystyle{unsrt}  

\section*{Acknowledgements}
The authors would like to acknowledge funding support from the National Science Foundation CRISP 2.0 Type 2 $\#$1832662. The authors would also like to acknowledge INRIX for providing the traffic data. Any opinions, findings, conclusions, or recommendations expressed in this research are those of the authors and do not necessarily reflect the view of the funding agencies.

\section*{References}
Abadi, A., Rajabioun, T., \& Ioannou, P. A. (2015). Traffic flow prediction for road transportation networks with limited traffic data. IEEE Transactions on Intelligent Transportation Systems, 16(2), 653-662.

Adeli, H., \& Jiang, X. (2008). Intelligent infrastructure: neural networks, wavelets, and chaos theory for intelligent transportation systems and smart structures. CRC Press.

Blumberg, A. F., Georgas, N., Yin, L., Herrington, T. O., \& Orton, P. M. (2015). Street-scale modeling of storm surge inundation along the New Jersey Hudson river waterfront. Journal of Atmospheric and Oceanic Technology, 32(8), 1486-1497.

Boto-Giralda, D., Díaz-Pernas, F. J., González-Ortega, D., Díez-Higuera, J. F., Antón-Rodríguez, M., Martínez-Zarzuela, M., \& Torre-Díez, I. (2010). Wavelet-based denoising for traffic volume time series forecasting with self-organizing neural networks. Computer-Aided Civil and Infrastructure Engineering, 25(7), 530-545.

Brown, S., \& Dawson, R. (2016). Building network-level resilience to resource disruption from flooding: Case studies from the Shetland Islands and Hurricane Sandy. In E3S Web of Conferences (Vol. 7, p. 04008). EDP Sciences.

Chang, C. H., Chung, M. K., Yang, S. Y., Hsu, C. T., \& Wu, S. J. (2018). A case study for the application of an operational two-dimensional real-time flooding forecasting system and smart water level gauges on roads in Tainan City, Taiwan. Water, 10(5), 574.

Chen, Y., Wang, Q., \& Ji, W. (2020). Rapid assessment of disaster impacts on highways using social media. Journal of Management in Engineering, 36(5), 04020068.

City of Houston. (2021). Code of Ordinances-Supplement 88 Update 1. 

Defferrard, M., Bresson, X., \& Vandergheynst, P. (2016). Convolutional neural networks on graphs with fast localized spectral filtering. In Advances in neural information processing systems (pp. 3844-3852).

Drobot, S. D., Benight, C., \& Gruntfest, E. C. (2007). Risk factors for driving into flooded roads. Environmental Hazards, 7(3), 227-234.

Dong, S., Yu, T., Farahmand, H., \& Mostafavi, A. (2020a). Probabilistic modeling of cascading failure risk in interdependent channel and road networks in urban flooding. Sustainable Cities and Society, 62, 102398.

Dong, S., Esmalian, A., Farahmand, H., \& Mostafavi, A. (2020b). An integrated physicalsocial analysis of disrupted access to critical facilities and community service-loss tolerance in urban flooding. Computers, Environment and Urban Systems, 80, 101443.

Dong, S., Yu, T., Farahmand, H., \& Mostafavi, A. (2020c). Bayesian modeling of flood control networks for failure cascade characterization and vulnerability assessment. Computer‐Aided Civil and Infrastructure Engineering, 35(7), 668-684.

Fan, C., Jiang, X., \& Mostafavi, A. (2020). A Network Percolation-based Contagion Model of Flood Propagation and Recession in Urban Road Networks. arXiv preprint arXiv:2004.03552.

Fan, C., \& Mostafavi, A. (2019). A graph‐based method for social sensing of infrastructure disruptions in disasters. Computer‐Aided Civil and Infrastructure Engineering, 34(12), 1055-1070.

FitzGerald, G., Du, W., Jamal, A., Clark, M., \& Hou, X. Y. (2010). Flood fatalities in contemporary Australia (1997–2008). Emergency Medicine Australasia, 22(2), 180-186.

Gauthier, P., Furno, A., \& El Faouzi, N. E. (2018). Road network resilience: how to identify critical links subject to day-to-day disruptions. Transportation research record, 2672(1), 54-65.

Hammond, D. K., Vandergheynst, P., \& Gribonval, R. (2011). Wavelets on graphs via spectral graph theory. Applied and Computational Harmonic Analysis, 30(2), 129-150.

Haworth, J., Shawe-Taylor, J., Cheng, T., \& Wang, J. (2014). Local online kernel ridge regression for forecasting of urban travel times. Transportation Research Part C: Emerging Technologies, 46, 151-178.

Helderop, E., \& Grubesic, T. H. (2019). Flood evacuation and rescue: The identification of critical road segments using whole-landscape features. Transportation research interdisciplinary perspectives, 3, 100022.

Jongman, B. (2018). Effective adaptation to rising flood risk. Nature communications, 9(1), 1-3.

Jonkman, S.N., Kelman, I., 2005. An analysis of the causes and circumstances of flood disaster deaths. Disasters, 29 (1), 75–97.

Li, Y., \& Yuan, Y. (2017). Convergence analysis of two-layer neural networks with relu activation. arXiv preprint arXiv:1705.09886.

Lu, H., Huang, D., Song, Y., Jiang, D., Zhou, T., \& Qin, J. (2020). St-trafficnet: A spatial-temporal deep learning network for traffic forecasting. Electronics, 9(9), 1474.

Lv, Y., Duan, Y., Kang, W., Li, Z., \& Wang, F.-Y. (2015). Traffic flow prediction with big data: A deep learning approach. IEEE Transactions on Intelligent Transportation Systems, 16(2), 865-873.

Manley, E., Cheng, T., Penn, A., \& Emmonds, A. (2014). A framework for simulating large-scale complex urban traffic dynamics through hybrid agent-based modelling. Computers, Environment and Urban Systems, 44, 27-36.

Min, X., Hu, J., Chen, Q., Zhang, T., \& Zhang, Y. (2009, October). Short-term traffic flow forecasting of urban network based on dynamic STARIMA model. In 2009 12th International IEEE conference on intelligent transportation systems (pp. 1-6). IEEE.

Naulin, J. P., Payrastre, O., \& Gaume, E. (2013). Spatially distributed flood forecasting in flash flood prone areas: Application to road network supervision in Southern France. Journal of Hydrology, 486, 88-99.

Niepert, M., Ahmed, M., \& Kutzkov, K. (2016, June). Learning convolutional neural networks for graphs. In International conference on machine learning (pp. 2014-2023).

Podesta, C., Coleman, N., Esmalian, A., Yuan, F. \& Mostafavi, A. (2020). Quantifying Community Resilience Based on Fluctuations in Visits to Point-of-Interest from Digital Trace Data. arXiv preprint arXiv:2011.07440.

Polson, N. G., and Sokolov, V. O. (2017). Deep learning for short-term traffic flow prediction. Transportation Research Part C: Emerging Technologies, 79, 1–17.

Pregnolato, M., Ford, A., Wilkinson, S. M., \& Dawson, R. J. (2017). The impact of flooding on road transport: A depth-disruption function. Transportation research part D: transport and environment, 55, 67-81.

Schnebele, E., Cervone, G., \& Waters, N. (2014). Road assessment after flood events using non-authoritative data. Natural Hazards and Earth System Sciences, 14(4), 1007-1015.

Versini, P. A., Gaume, E., and Andrieu, H. (2010). Application of a distributed hydrological model to the design of a road inundation warning system for flash flood prone areas. Natural Hazards \& Earth System Sciences, 10(4), 805-817.

Wang, W., Yang, S., Stanley, H. E., \& Gao, J. (2019). Local floods induce large-scale abrupt failures of road networks. Nature communications, 10(1), 1-11.

Yang, H. F., Dillon, T. S., \& Chen, Y. P. P. (2016). Optimized structure of the traffic flow forecasting model with a deep learning approach. IEEE transactions on neural networks and learning systems, 28(10), 2371-2381.

Yu, B., Yin, H., \& Zhu, Z. (2018). Spatio-temporal graph convolutional networks: a deep learning framework for traffic forecasting. In Proceedings of the 27th International Joint Conference on Artificial Intelligence (pp. 3634-3640).

Yin, J., Yu, D., \& Wilby, R. (2016). Modelling the impact of land subsidence on urban pluvial flooding: A case study of downtown Shanghai, China. Science of the Total Environment, 544, 744-753.

Yin, J., Yu, D., Yin, Z., Liu, M., \& He, Q. (2016). Evaluating the impact and risk of pluvial flash flood on intra-urban road network: A case study in the city center of Shanghai, China. Journal of hydrology, 537, 138-145.

Yu, R., Li, Y., Shahabi, C., Demiryurek, U., \& Liu, Y. (2017). Deep learning: A generic approach for extreme condition traffic forecasting. In Proceedings of the 2017 SIAM international Conference on Data Mining (pp. 777-785). Society for Industrial and Applied Mathematics.

Yuan, F., Liu, R., Mao, L., \& Li, M. (2021a). Internet of people enabled framework for evaluating performance loss and resilience of urban critical infrastructures. Safety Science, 134, 105079.

Yuan, F., Li, M., Liu, R., Zhai, W., \& Qi, B. (2021b). Social media for enhanced understanding of disaster resilience during Hurricane Florence. International Journal of Information Management, 57, 102289.

Yuan, F., Esmalian, A., Oztekin, B., \& Mostafavi, A. (2021c). Unveiling Spatial Patterns of Disaster Impacts and Recovery Using Credit Card Transaction Variances. arXiv preprint arXiv:2101.10090.

Yuan, F., \& Liu, R. (2020). Mining social media data for rapid damage assessment during Hurricane Matthew: feasibility study. Journal of Computing in Civil Engineering, 34(3), 05020001.

Yuan, F., Li, M., \& Liu, R. (2020a). Understanding the evolutions of public responses using social media: Hurricane Matthew case study. International Journal of Disaster Risk Reduction, 51, 101798.

Yuan, F., Li, M., Zhai, W., Qi, B., \& Liu, R. (2020b). Social Media Based Demographics Analysis for Understanding Disaster Response Disparity. In Construction Research Congress 2020: Computer Applications (pp. 1020-1028). Reston, VA: American Society of Civil Engineers.

Yuan, F., \& Liu, R. (2019). Identifying damage-related social media data during Hurricane Matthew: A machine learning approach. In Computing in Civil Engineering 2019: Visualization, Information Modeling, and Simulation (pp. 207-214). Reston, VA: American Society of Civil Engineers.

Yuan, F., \& Liu, R. (2018a). Feasibility study of using crowdsourcing to identify critical affected areas for rapid damage assessment: Hurricane Matthew case study. International journal of disaster risk reduction, 28, 758-767.

Yuan, F., \& Liu, R. (2018b). Integration of social media and unmanned aerial vehicles (UAVs) for rapid damage assessment in Hurricane Matthew. In Construction Research Congress 2018 (pp. 513-523).

Yuan, F., \& Liu, R. (2018c). Crowdsourcing for forensic disaster investigations: Hurricane Harvey case study. Natural Hazards, 93(3), 1529-1546.

Zhai, W., Peng, Z. R., \& Yuan, F. (2020). Examine the effects of neighborhood equity on disaster situational awareness: Harness machine learning and geotagged Twitter data. International Journal of Disaster Risk Reduction, 101611.

Zhang, C., Yao, W., Yang, Y., Huang, R., \& Mostafavi, A. (2020). Semiautomated social media analytics for sensing societal impacts due to community disruptions during disasters. Computer‐Aided Civil and Infrastructure Engineering.

Zhang, Y., Cheng, T., \& Ren, Y. (2019). A graph deep learning method for short‐term traffic forecasting on large road networks. Computer‐Aided Civil and Infrastructure Engineering, 34(10), 877-896.

Zhang, J., Zheng, Y., Qi, D., Li, R.,\& Yi, X. (2016). DNN-based prediction model for spatio-temporal data. In Proceedings of the 24th ACM SIGSPATIAL International Conference on Advances in Geographic Information Systems (pp. 1-4).

\end{document}